\documentclass{article}

\usepackage{microtype}
\usepackage{graphicx}
\usepackage{subcaption}
\usepackage{booktabs} 

\usepackage{hyperref}

\usepackage[preprint]{icml2026}

\usepackage{amsmath}
\usepackage{amssymb}
\usepackage{mathtools}
\usepackage{amsthm}
\usepackage{hyperref}       
\usepackage{url}            
\usepackage{booktabs}       
\usepackage{amsfonts}       
\usepackage{graphicx}       
\usepackage{nicefrac}       
\usepackage{microtype}      
\usepackage{xcolor}         
\usepackage{enumitem}
\usepackage{amsmath}
\usepackage{pifont}
\usepackage{makecell}

\usepackage{xcolor}
\usepackage[most]{tcolorbox}
\usepackage{amsmath}     
\usepackage{amssymb}     
\usepackage{needspace}
\usepackage{afterpage}
\usepackage{wrapfig}
\usepackage{bm} 
\usepackage{multirow}
\usepackage{subcaption}
\usepackage{graphicx}
\usepackage{xcolor}
\usepackage[normalem]{ulem}
\usepackage{colortbl}
\usepackage{booktabs}
\usepackage{mathtools}
\usepackage{subcaption}
\usepackage{booktabs}   
\usepackage{multirow}  
\usepackage{amsfonts}
\usepackage[table]{xcolor}

\usepackage{url}
\usepackage[capitalize,noabbrev]{cleveref}

\theoremstyle{plain}

\theoremstyle{definition}

\theoremstyle{remark}

\def\Algnameabbr{Stable-OPD}

\usepackage[textsize=tiny]{todonotes}

\icmltitlerunning{Demystifying OPD: Length Inflation and Stabilization Strategies for Large Language Models}

\newcommand{\rionedistill}{${\dagger}$}      
\newcommand{\openthinker}{${\ddagger}$}      

\definecolor{backbonegray}{gray}{0.93}

\begin{document}

\twocolumn[
  \icmltitle{Demystifying OPD: Length Inflation and \\ Stabilization Strategies for Large Language Models}



  \icmlsetsymbol{equal}{*}

  \begin{icmlauthorlist}
    \icmlauthor{Feng Luo}{yyy}
    \icmlauthor{Yu-Neng Chuang}{yyy}
    \icmlauthor{Guanchu Wang}{comp}
    \icmlauthor{Zicheng Xu}{sch}\\
    \icmlauthor{Xiaotian Han}{zzz}
    \icmlauthor{Tianyi Zhang}{yyy}
    \icmlauthor{Vladimir Braverman}{sch}
  \end{icmlauthorlist}

  \icmlaffiliation{yyy}{Department of Computer Science, Rice University, Houston, USA}
  \icmlaffiliation{comp}{Department of Computer Science, University of North Carolina at Charlotte, Charlotte, USA}
  \icmlaffiliation{sch}{Department of Computer Science, Johns Hopkins University, Baltimore, USA}
  \icmlaffiliation{zzz}{Department of Computer and Data Sciences, Case Western Reserve University}
  \icmlcorrespondingauthor{Feng Luo}{fl38@rice.edu}
  \icmlcorrespondingauthor{Yu-Neng Chuang}{yc146@rice.edu}

  \icmlkeywords{}

  \vskip 0.3in
]



\printAffiliationsAndNotice{}  

\begin{abstract}
  On-policy distillation (OPD) trains student models under their own induced distribution while leveraging supervision from stronger teachers. We identify a failure mode of OPD: as training progresses, on-policy rollouts can undergo abrupt length inflation, causing truncated trajectories to dominate the training data. This truncation collapse coincides with abrupt repetition saturation and induces biased gradient signals, leading to severe training instability and sharp degradation in validation performance. We attribute this problem to the interaction between student-induced data collection and the distillation objective, which implicitly favors long and repetitive rollouts. To address this issue, we propose \Algnameabbr{}, a stabilized OPD framework that combines a reference-based divergence constraint with rollout mixture distillation. These together mitigate repetition-induced length inflation and further stabilize OPD training.
  Across multiple math reasoning datasets, our approach prevents truncation collapse, stabilizes training dynamics, and improves performance by 7.2\% on average.
\end{abstract}

\section{Introduction}
On-policy distillation (OPD)~\cite{lai2020dual,czarnecki2019distilling} has recently gained attention as an effective framework for training student LLMs under their own induced distribution while leveraging supervision from stronger teacher LLMs. By iteratively collecting rollouts from the student policy and applying distillation losses, OPD avoids the distribution mismatch inherent to purely offline distillation and enables continual adaptation during training~\cite{agarwal2024policy, lu2025onpolicydistillation, yang2025qwen3, ye2025black}. This paradigm has shown promise in domains such as long-form generation and reasoning, where robustness under the student’s evolving policy is critical.

\begin{figure}
  \centering
  \includegraphics[width=1\linewidth]{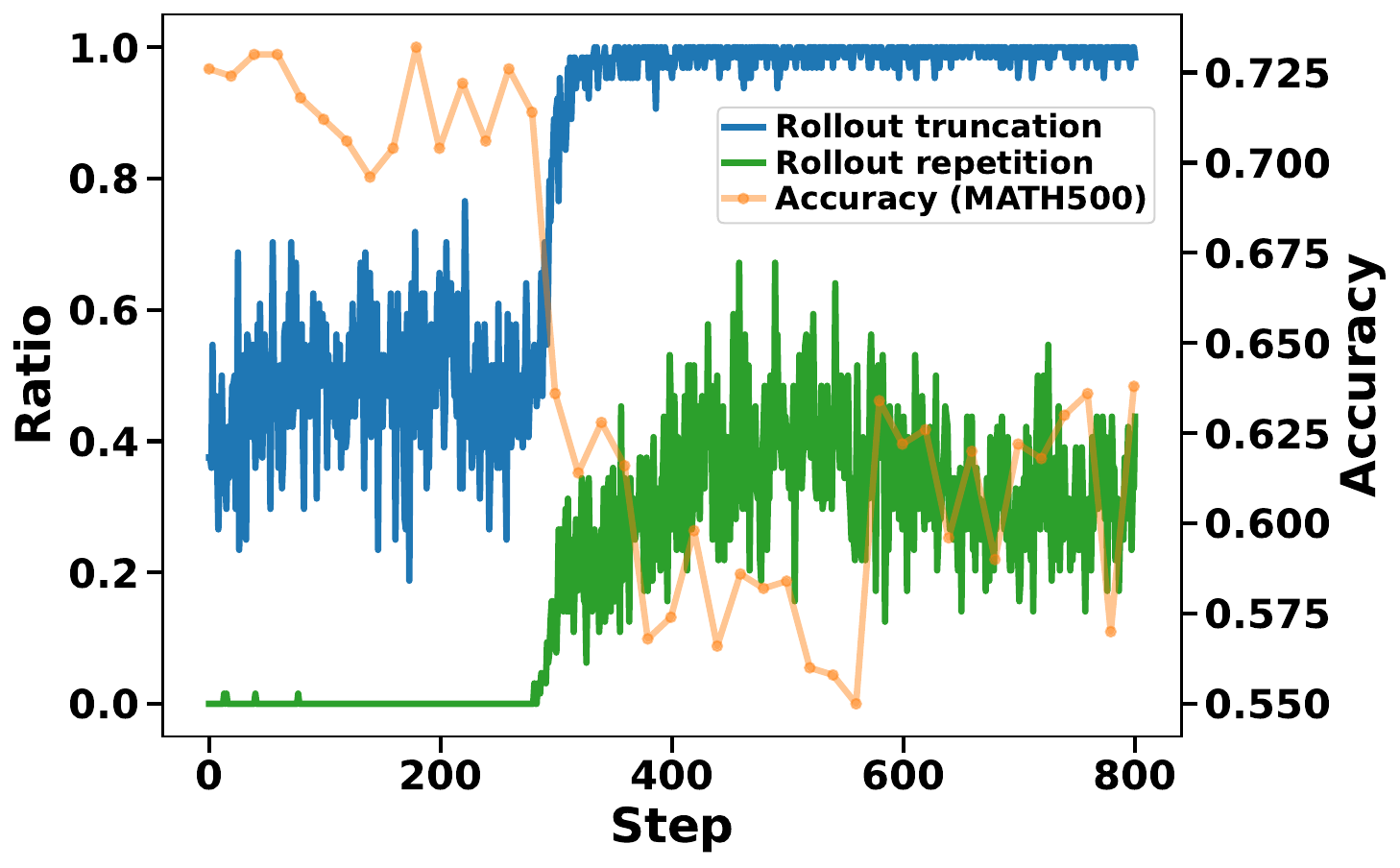}
  \vspace{-0.6cm}
  \caption{Abrupt length inflation within OPD.}
  \label{fig:demo}
  \vspace{-0.3cm}
\end{figure}

Despite the strong performance of OPD, we identify a training pathology that emerges during optimization. As depicted in Fig.~\ref{fig:demo}, abrupt rollout length inflation occurs as training progresses: student-generated rollouts suddenly grow much longer, causing truncated trajectories terminated by a fixed context or token limit to dominate the training data.
We show that this behavior is not a generic form of length bias commonly observed in GRPO-style RL training. Instead, it is driven by a primary failure mode specific to OPD, which we term ``abrupt repetition saturation". In this regime, student rollouts unexpectedly enter repetitive generation patterns, where repeated tokens rapidly dominate the generated sequence. Once repetition saturation occurs, subsequent on-policy updates reinforce this behavior, leading to rollout length inflation and eventual training collapse.

The underlying mechanism lies in OPD’s optimization under the student-induced distribution, guided by reverse-KL advantage signals derived from teacher-student likelihood discrepancy. During abrupt repetition saturation, repetitive tokens receive systematically larger reverse-KL advantages than regular tokens. While these tokens contribute little when they are rare, their frequency rises sharply once repetition begins, and their disproportionately large advantages then make them increasingly dominant in the gradient update. This creates a self-reinforcing feedback loop: updates increasingly favor repetitive continuations, which further encourages repetition and length expansion. In this sense, student repetition may exploit or hack the teacher's likelihood-based signal.

Empirically, shown in Sec.~\ref{sec:understand-inflation}, we observe that the onset of repetition saturation and the rise in rollout truncation tightly coincide with severe training instability, including sudden drops in validation accuracy and biased learning gradients. Importantly, this collapse occurs while the teacher model and loss formulation remain fixed, indicating that the instability is intrinsic to OPD's on-policy dynamics under abrupt truncation and repetition. Together, these observations suggest that truncation-dominated OPD training yields biased gradient signals that destabilize optimization.

To address this training challenge, we introduce a unified stabilization framework, \Algnameabbr{}, which stabilizes OPD through two complementary mechanisms. First, we incorporate a reference-based divergence constraint that limits uncontrolled policy drift and curbs excessive rollout expansion. Second, we employ rollout mixture distillation, which blends on-policy student rollouts with reference trajectories to maintain a stable fraction of complete, non-truncated sequences throughout training. 
Across multiple mathematical reasoning benchmarks, we show that \Algnameabbr{} consistently stabilizes OPD training and improves average accuracy by 7.2\% compared to standard OPD baselines. Our contributions are summarized as follows:
\vspace{-2mm}
\begin{itemize}[leftmargin=10pt]
    \itemsep=-1pt
    \item \textbf{Length Inflation:} We identify rollout length inflation as an observable training pathology in OPD with a primary failure mode of \emph{abrupt repetition saturation}.
    \item \textbf{Rollout Pathology:} We empirically show that repetition-saturated, truncation-dominated rollouts produce biased gradients that destabilize training.
    \item \textbf{Stabilization Protocol:} We identify the specific constraints required to stop this pathology. We show that a dual strategy of divergence constraint and rollout mixture is effective to prevent the student from hacking the distillation objective.
    \item \textbf{Evaluation:} Across six datasets and three LLMs, \Algnameabbr{} consistently improves accuracy and reduces repetition saturation.
\end{itemize}

\section{Preliminary}
\label{sec:premliminary}
\subsection{Group Relative Policy Optimization (GRPO)}
Group Relative Policy Optimization (GRPO) is a reinforcement learning method commonly used in RL with verifiable rewards (RLVR) for tasks such as math and code \cite{shao2024deepseekmath,guo2025deepseek,zhang2025survey}. 
Given a prompt $x$, GRPO samples a group of $G$ responses $\{o_i\}_{i=1}^G$ from policy $\pi_{\theta_{\text{old}}}$ and assigns each response a
\emph{sequence-level} reward $r_i = R(x,o_i)$ , which is often binary for
verifiable tasks (e.g., correctness). This method then assigns a group-normalized advantage $A_i$ to all tokens $k=1, ... |o_{i}|$ within response $o_{i}$.
Then, the objective is inherited from the clipped objective proposed by PPO\cite{schulman2017proximal},
\begin{equation}
\begin{aligned}
&\mathcal{J}_{\mathrm{GRPO}}(\theta)
= \mathbb{E}_{x\sim P(X), \{o_i\}_{i=1}^{G}\sim \pi_{\theta_{\rm old}}}
\\&\frac{1}{G}\sum_{i=1}^G \frac{1}{|o_i|}
\sum_{t=1}^{|o_i|}
\min\big(\rho_i^t(\theta)A_i,\, 
\mathrm{clip}(\rho_i^t(\theta),1-\epsilon,1+\epsilon)A_i\big)
.
\end{aligned}
\label{eq:grpo-obj}
\end{equation}
where $P(X)$ refers to the question distribution, $\rho_i^t(\theta)=\pi_\theta(o_i^t|x,o_i^{<t})/\pi_{\theta_{\text{old}}}(o_i^t|x,o_i^{<t})$, $\pi_\theta$ is the current policy, and $\epsilon$ controls the trust region of policy updates. 
Despite strong empirical performance, GRPO has two notable limitations. First, the reward signal is sparse and provided only at the sequence level, offering limited token-level guidance on where the model makes mistakes. Second, when sampled responses within a group are all correct or all incorrect, their advantages become zero, yielding no effective update despite the computational cost of group sampling. In this work, we leverage the clipping objective of GRPO with token-level advantages for OPD training.


\subsection{Knowledge distillation for LLM}
Knowledge distillation\cite{hinton2015distilling,rusu2015policy,kim2016sequence,gou2021knowledge} trains a student model to learn from a more capable teacher by matching the
teacher’s output distribution. Standard knowledge distillation typically trains the student on a fixed set of sequences, such as teacher-generated responses or ground-truth demonstrations. 
Let the student have learnable parameters $\theta$, with $\pi_S^\theta$
differentiable with respect to $\theta$. 
Given a fixed dataset of input-output sequence pairs $(X,Y)$ and a divergence $D$, standard distillation minimizes the expected discrepancy between teacher and student next-token distributions $\pi_T$ and $\pi_S^\theta$ along the fixed sequences: $\mathcal{L}_{\mathrm{SD}}(\theta) = \mathbb{E}_{(x,y)\sim P(X,Y)}
\Big[
D\!\big(
\pi_T \,\|\, \pi_S^\theta
\big)(y \mid x)
\Big]$.
While effective, this off-policy training paradigm introduces a training-inference mismatch: during inference, the student conditions on its own generated prefixes, which may deviate from the prefixes observed in the fixed distillation dataset. OPD\cite{gu2023minillm,agarwal2024policy,lu2025onpolicydistillation} addresses this issue by training on student-generated responses $\hat{y}\sim \pi_S^\theta$, thereby aligning  the training states with the student’s test-time states. The loss for OPD is:
\begin{equation}
\mathcal{L}_{\mathrm{OPD}}(\theta)
\;=\;
\mathbb{E}_{x \sim P(X),\hat{y}\sim \pi_S^\theta}
\Big[
D\!\big(
\pi_S^\theta \,\|\, \pi_T
\big)(\hat{y} \mid x)
\Big].
\label{eq:onpolicy_distill}
\end{equation}

\section{Length Inflation in OPD}
\label{sec:opd-failure}
In this section, we formulate OPD for reasoning tasks, define metrics for probing training dynamics, present the empirical failure mode of abrupt truncation-repetition inflation, and analyze it from rollout-level, token-level, and mechanistic perspectives.

\subsection{OPD Training for LLM Reasoning}
\label{sec:opd-training}
\textbf{Reverse KL reward and token-level advantage.}
Instead of sequence-level rewards $r_i$, we use a teacher model $\pi_T$ to define a token-level reward on each visited state. Following prior work~\cite{lu2025onpolicydistillation}, we define for each student token $y_{i,t}$ the reverse-KL-based reward:
\begin{equation}
\notag r_{i,t}^{\mathrm{KL}}
\;=\;
\log \pi_T(\hat y_{i,t} \mid s_{i,t}, \hat y_{i, <t})
-
\log \pi_\theta(\hat y_{i,t} \mid s_{i,t}, \hat y_{i, <t}),
\label{eq:kl_reward}
\end{equation}
which encourages the student to increase the probability of tokens to which the teacher assigns high likelihood.
We then take the token-level advantage to be $A_{i,t} \triangleq r_{i,t}^{\mathrm{KL}}$. This token-level advantage contrasts with GRPO, where a single sequence-level advantage $A_i$ is broadcast to all tokens in a response $o_i$.

\textbf{Objective.}
The overall optimization objective keeps the GRPO-style clipped form of
Eq.~\eqref{eq:grpo-obj}, but replaces the sequence-level advantage $A_i$
with the token-level advantages $A_{i,t}$.
This GRPO-style OPD objective provides dense token-level guidance from the
teacher and serves as our default training setup in this work.

\subsection{Metrics for Analyzing OPD Dynamics}
To study the dynamics of OPD training, we monitor two simple metrics computed
over a set of model rollouts: \emph{truncation rate} and \emph{repetition rate}.
Both metrics can be evaluated on on-policy training rollouts and on held-out
validation prompts; for clarity we define them on an arbitrary set of
student-generated responses $R = \{o_i\}_{i=1}^N$.

\textbf{Truncation rate.}
Each rollout $o_i$ is generated under a fixed maximum generation length.
We say $o_i$ is \emph{truncated} if generation terminates because this length
budget is exhausted, rather than because the model emits an EOS token.
Let $\mathrm{trunc}(o_i)\in\{0,1\}$ indicate whether $o_i$ is truncated.
The truncation rate over $R$ is the average value of $\mathrm{TruncRate} = \frac{1}{N}\sum_{i=1}^{N} \mathrm{trunc}(o_i)(R)$.

\textbf{Repetition rate.}
To capture degenerate generations with strong local repetition, we use a
compression-based repetition metric.
For a rollout $o_i$, let $\mathrm{bytes}(o_i)$ denote its byte representation,
and let $c(\cdot)$ denote zlib compression with a fixed level.
We compute the compression ratio as $\mathrm{CompRatio}(o_i) = \big\lvert \mathrm{bytes}(o_i^{\text{tail}}) \big\rvert/\big\lvert c\big(\mathrm{bytes}(o_i^{\text{tail}})\big) \big\rvert$,
where $o_i^{\text{tail}}$ is the suffix consisting of the last $L$ characters
of $o_i$.
We say $o_i$ exhibits high repetition if it is sufficiently long and its tail
is highly compressible:
$
\mathrm{rep}(o_i)
=
\mathbf{1}\!\left[
\lvert o_i^{\text{tail}} \rvert > L
\;\land\;
\mathrm{CompRatio}(o_i) > \tau
\right]
$,
where we set $L=10{,}000$ and $\tau=10$ in our experiments.
The repetition rate over $R$ is then defined as the average value of $\mathrm{RepRate}(R)=\frac{1}{N}\sum_{i=1}^{N} \mathrm{rep}(o_i)$.

In our implementation, $\mathrm{RepRate}$  measures the fraction of
long rollouts whose tails exhibit extreme compressibility, which correlates
well with visibly repetitive, low-information continuations.

\subsection{Empirical Failure Mode: Abrupt Truncation-Repetition Inflation}
\label{sec:opd-empirical-failure-mode}
\begin{figure*}[t]
    \centering

    \begin{subfigure}[t]{1.0\textwidth}
        \centering
        \includegraphics[width=\linewidth]{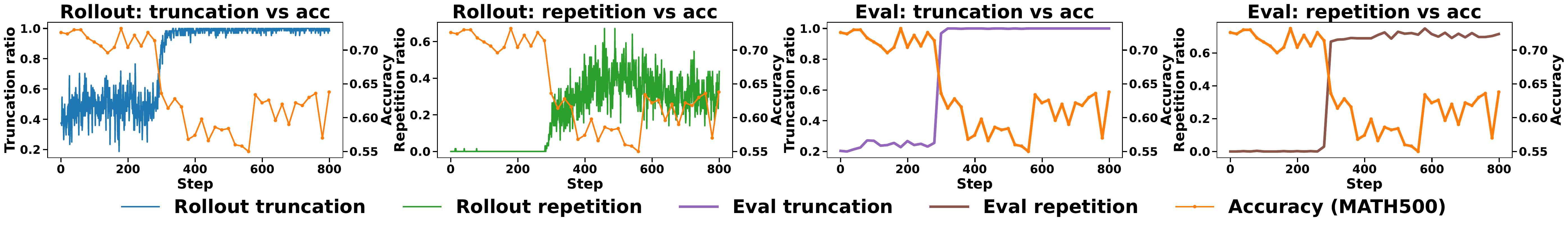}
        \vspace{-6mm}
        \caption{\footnotesize Student: Qwen2.5-Math-1.5B; Teacher: DeepSeek-R1-Distill-7B}
        \vspace{1mm}
    \end{subfigure}
    \begin{subfigure}[t]{1.0\textwidth}
        \centering
        \includegraphics[width=\linewidth]{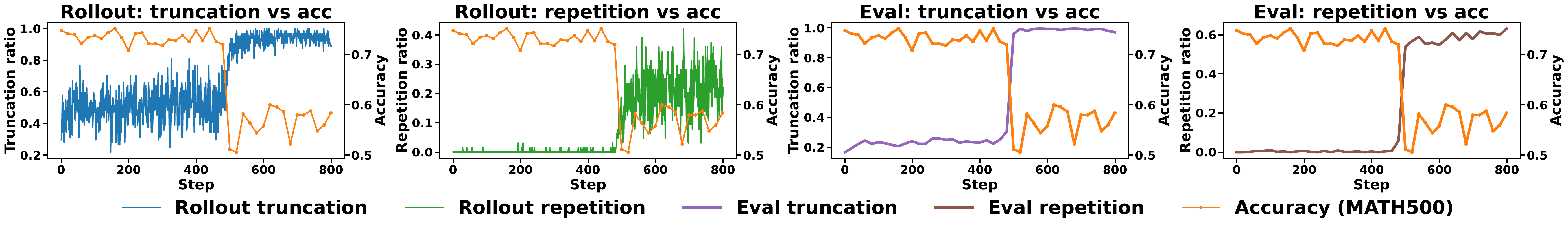}
        \vspace{-6mm}
        \caption{\footnotesize Student: Qwen2.5-Math-1.5B; Teacher: OpenThinker3-7B}
        \vspace{1mm}
    \end{subfigure}
    \begin{subfigure}[t]{1.0\textwidth}
        \centering
        \includegraphics[width=\linewidth]{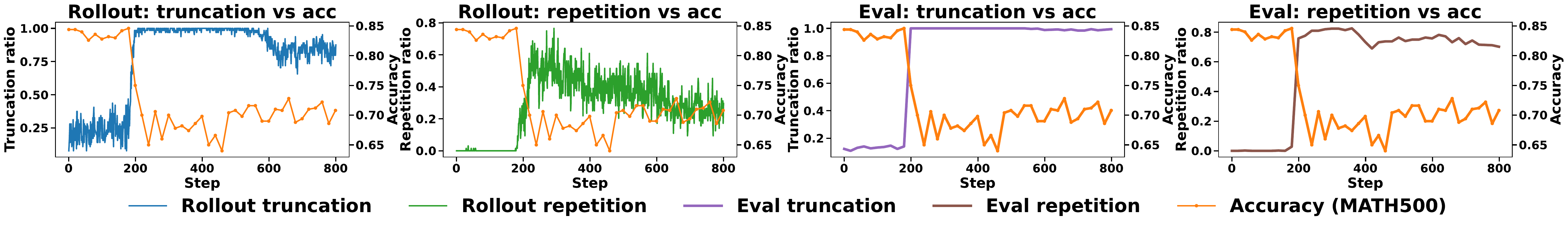}
        \vspace{-6mm}
        \caption{\footnotesize Student: Qwen2.5-Math-7B; Teacher: DeepSeek-R1-Distill-7B}
        \vspace{1mm}
    \end{subfigure}
    \caption{
    Training dynamics of OPD on three groups. Training starts in a stable regime with low truncation and repetition, followed by a sharp phase transition where truncation and repetition increase and remain high while validation accuracy collapses, illustrating a robust truncation-repetition inflation failure mode of OPD.
    }
    \label{fig:trunc_rep_acc_panels}
\end{figure*}

We investigate OPD dynamics on a 13k subset of the OpenR1-Math-220k reasoning data and consider three
student-teacher groups that vary both student scale and
teacher choice: (i) Qwen2.5-Math-1.5B student with DeepSeek-R1-Distill-7B
teacher, (ii) Qwen2.5-Math-7B student with OpenThinker3-7B teacher, and
(iii) Qwen2.5-Math-7B student with DeepSeek-R1-Distill-7B teacher.
For each configuration, we run OPD with the reward setting in
Sec.~\ref{sec:opd-training} and track the truncation and repetition metrics on both training rollouts and MATH500~\cite{lightman2023lets} validation set, together with validation accuracy. The resulting dynamics are shown in Fig.~\ref{fig:trunc_rep_acc_panels}.

\textbf{Stable early training stage.}
Across all three settings, OPD initially behaves as desired. Validation accuracy gradually improves, most student responses finish within the generation budget, and visibly repetitive tails are rare. The rollout truncation rate $\mathrm{TruncRate}$ stays around $0.5$ for Qwen2.5-Math-1.5B and around $0.23$ for Qwen2.5-Math-7B, while the validation truncation rate remains near $0.2$ and $0.1$, respectively. The repetition rate $\mathrm{RepRate}$ stays close to zero on both training and validation generations.

\textbf{Phase transition and robustness.} As training progresses, all three settings exhibit a sharp phase transition. Within a relatively short window of OPD steps (about $30$ steps in Fig.~\ref{fig:trunc_rep_acc_panels}), the truncation rate on on-policy rollouts rises abruptly toward one, indicating that most generations now hit the maximum length budget without emitting an EOS token. At the same time, the repetition rate $\mathrm{RepRate}$ spikes from near zero to about $0.3$-$0.6$, revealing the emergence of long, highly compressible suffixes dominated by repetitive patterns. 

The same qualitative transition appears on the MATH500 validation set: both validation $\mathrm{TruncRate}$ and $\mathrm{RepRate}$ jump sharply at nearly the same training step, coinciding with a sudden drop in validation accuracy. 
Observing this truncation-repetition inflation and worse validation performance across all three student-teacher groups suggests that it is a robust OPD failure mode in reasoning, rather than an artifact of a particular model pair or dataset split.
We refer to this phenomenon as \emph{abrupt truncation-repetition inflation}.

\subsection{Understanding Abrupt Repetition Inflation}
\label{sec:understand-inflation}
\begin{figure}[t]
  \centering
  \begin{subfigure}[t]{0.5\textwidth}
      \centering
      \includegraphics[width=\linewidth]{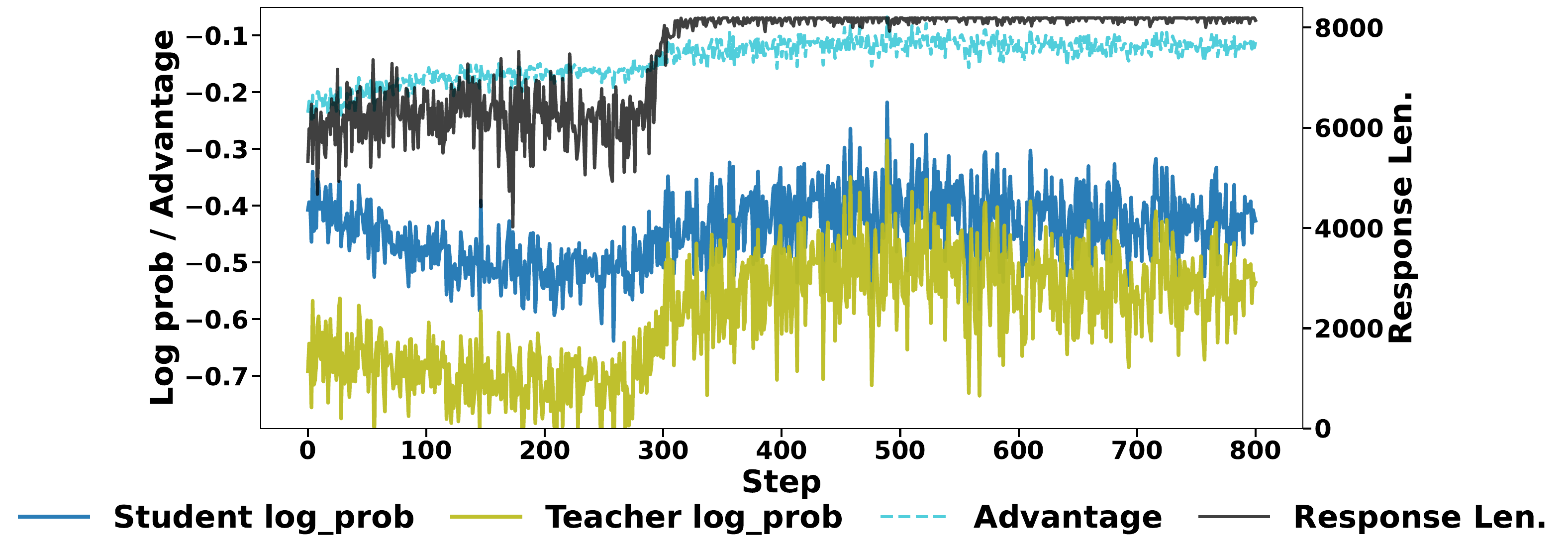}
      \vspace{-6mm}
      \caption{\footnotesize Student: Qwen2.5-Math-1.5B; Teacher: DeepSeek-R1-Distill-7B}
      \vspace{1mm}
  \end{subfigure}
  \begin{subfigure}[t]{0.5\textwidth}
      \centering
      \includegraphics[width=\linewidth]{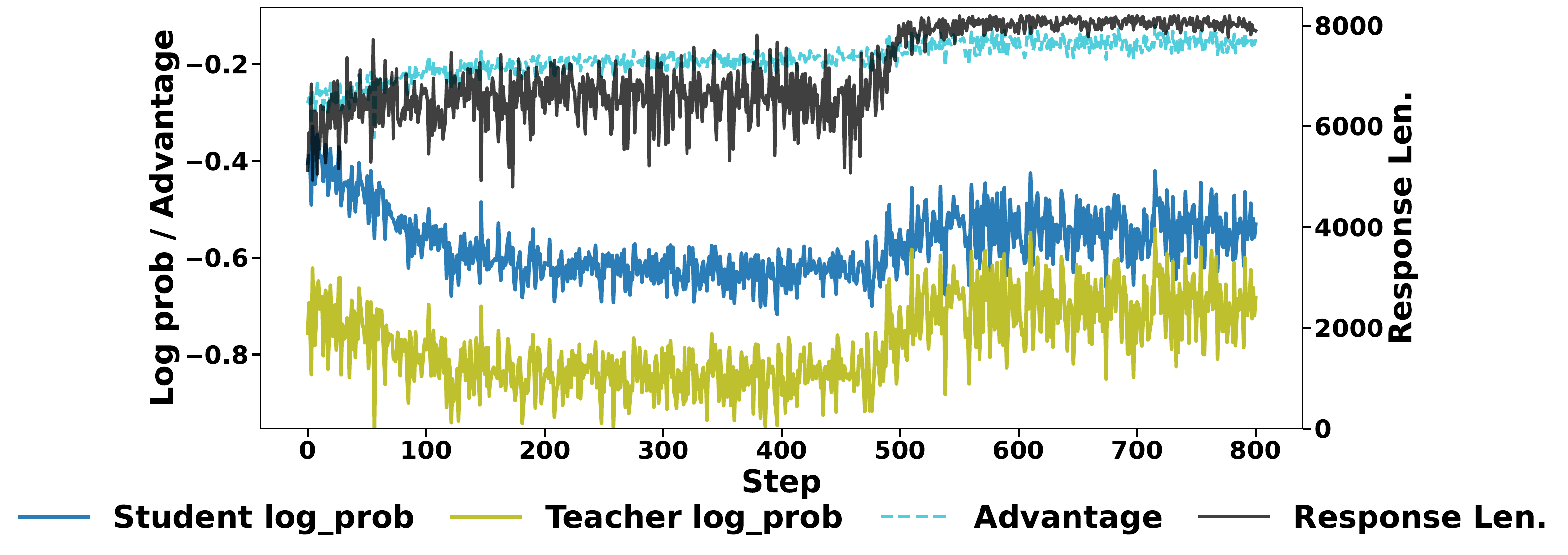}
      \vspace{-6mm}
      \caption{\footnotesize Student: Qwen2.5-Math-1.5B; Teacher: OpenThinker3-7B}
      \vspace{1mm}
  \end{subfigure}
  \begin{subfigure}[t]{0.5\textwidth}
      \centering
      \includegraphics[width=\linewidth]{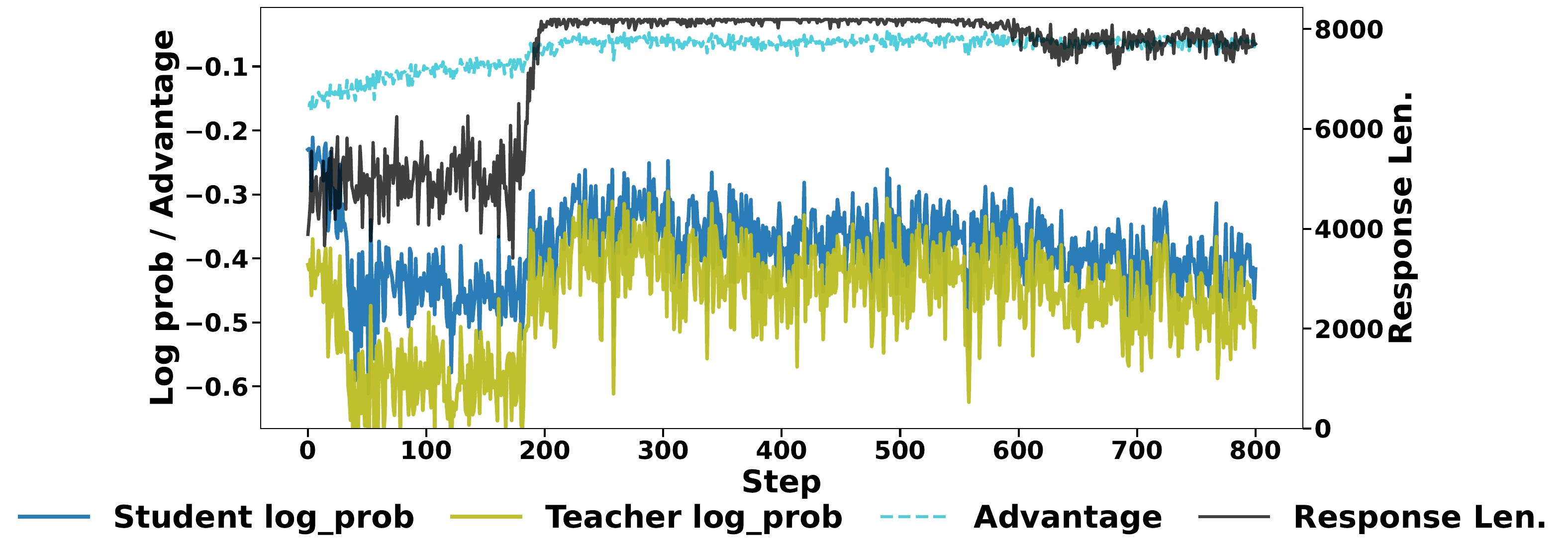}
      \vspace{-6mm}
      \caption{\footnotesize Student: Qwen2.5-Math-7B; Teacher: DeepSeek-R1-Distill-7B}
      \vspace{1mm}
  \end{subfigure}
  \caption{
    Rollout-level evidence of abrupt repetition inflation for three
    student-teacher groups. Around the step where rollout length abruptly inflates, both student and teacher log\_prob become much less negative, with the teacher's increase being larger, which induces a sudden jump in the reveser KL advantage.
  }
  \label{fig:advantage_len}
  \vspace{-0.3cm}
\end{figure}

\begin{figure*}[t]
    \centering
    \includegraphics[width=0.8\linewidth]{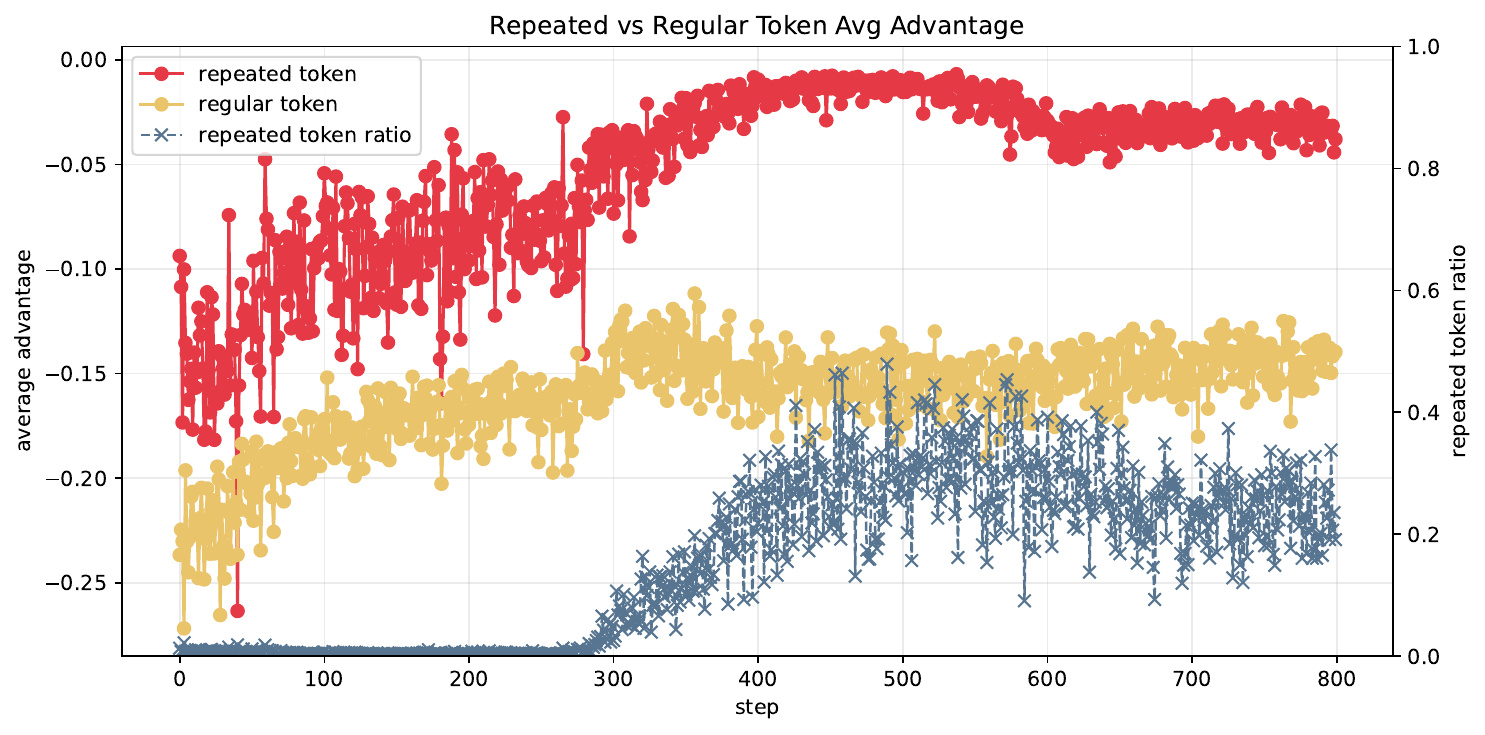}
    \caption{
    Reverse-KL advantage for regular and repetitive tokens during OPD training.  Repetitive tokens receive larger advantages than regular tokens throughout training.
    }
    \label{fig:token_advantage}
\end{figure*}

The previous subsection established the failure mode empirically. We now examine the training signals associated with its onset through rollout-level and token-level analyses.

\textbf{Rollout-level Evidence.} Fig.~\ref{fig:advantage_len} tracks the average student log-probability, teacher log-probability, reverse-KL advantage, and response length over training. In the early phase, response lengths remain moderate and all three statistics evolve smoothly. Around the onset of inflation, however, they shift together: response length jumps toward the generation budget, both log-probabilities become much less negative, and the teacher log-probability increases more than the student's, causing the average advantage to rise sharply. This pattern is consistent across all three student-teacher groups.

\textbf{Token-level Evidence.} To isolate the local reward signal, we compare the average reverse-KL advantage of regular tokens and repetitive tokens during training. As shown in Fig.~\ref{fig:token_advantage}, repetitive tokens receive larger advantages than regular tokens throughout training. Before the inflation phase, however, repetitive tokens are extremely rare and therefore contribute little to the aggregate update. Once inflation begins, their frequency rises sharply while their advantage remains larger. This provides direct empirical evidence that the reverse-KL signal is not uniformly distributed across the trajectory and systematically favors locally repetitive continuations.

Taken together, the rollout-level and token-level evidence suggests that OPD increasingly favors repetitive regions of the trajectory, especially 
once such tokens become more prevalent during training. We next formalize how favorable local reward and on-policy sampling jointly amplify this effect.

\subsection{Mechanistic Explanation of Abrupt Repetition Inflation}
\label{sec:mechanism-sketch}

To formalize the intuition above, we write the effective OPD update in a state-action form. For the ease of exposition, we ignore the clipping term and suppress the distinction between the rollout policy and the updated policy. 
Let $d_{\pi_\theta}(s)$ denote the state-visitation distribution induced by the current student policy over prefixes $s$, and let $A(s,y)=\log \pi_T(y\mid s)-\log \pi_\theta(y\mid s)$ denote the token-level reverse-KL advantage in Eq.~\eqref{eq:kl_reward}. Denoting the resulting policy gradient by $g(\theta)$, we have:
\begin{equation}
g(\theta)
\propto
\mathbb{E}_{s\sim d_{\pi_\theta},\, y\sim \pi_\theta(\cdot\mid s)}
\left[
A(s,y)\nabla_\theta \log \pi_\theta(y\mid s)
\right].
\label{eq:mechanism-gradient}
\end{equation}
Thus, the OPD update is governed by two coupled quantities: how often a state is visited, and how strongly actions at that state are favored by the reverse-KL signal. Let $\mathcal{R}$ denote the set of tokens in repetitive tails. We can decompose Eq.~\eqref{eq:mechanism-gradient} into contributions from states inside and outside $\mathcal{R}$:
\begin{equation}
g(\theta)
=
\mathbb{E}_{s\sim d_{\pi_\theta}}
\left[
\mathbf{1}\{s\notin \mathcal{R}\}\,\Delta(s)
\right]
+
\mathbb{E}_{s\sim d_{\pi_\theta}}
\left[
\mathbf{1}\{s\in \mathcal{R}\}\,\Delta(s)
\right],
\label{eq:mechanism-decompose}
\end{equation}
where
$\Delta(s)\triangleq
\mathbb{E}_{y\sim \pi_\theta(\cdot\mid s)}
\left[
A(s,y)\nabla_\theta \log \pi_\theta(y\mid s)
\right]$.

This decomposition makes the OPD-specific feedback explicit. Once visitation to $\mathcal{R}$ increases, the second term occupies a larger share of the update, and that update further encourages continuations that remain in $\mathcal{R}$. 
This creates a self-reinforcing loop: repetitive tails need not to be the most frequent tokens overall, but once they become sufficiently common, their combination of frequency and disproportionately large token-level advantages allows them to steer subsequent OPD updates toward repetitive continuations.

The empirical evidence in Fig.~\ref{fig:token_advantage} supports this pattern. Before collapse, repetitive tokens are relatively rare, so their total contribution to the update remains limited even if their average advantage is larger. However, around the transition their frequency rises sharply while their advantage remains much larger; in Fig.~\ref{fig:token_advantage}, repetitive tokens account for roughly $30\%$ of tokens after collapse, yet their average advantage is about $4$--$9\times$ that of non-repetitive tokens. As a result, their aggregate influence on the update can increase disproportionately.

\section{Mitigating Repetition Saturation}
As shown in Sec.~\ref{sec:opd-failure}, standard OPD can reach a stage where
training rollouts are dominated by long, repetitive, and truncated
trajectories, leading to an abrupt truncation-repetition collapse and a sharp
drop in validation accuracy.
In this section, we introduce two strategies designed to explicitly mitigate
this failure mode.

\subsection{Mixture Distillation: Combining On- and Off-Policy Supervision}
The training objective of OPD is driven by
student-generated rollouts, and there is no explicit control over the
distribution of states the student visits.
Once the student starts to  visit long, repetitive, and truncated
trajectories, OPD updates are dominated by these degenerate states, which
accelerates truncation-repetition saturation and accuracy collapse.

To address this, we introduce a hybrid training paradigm, mixture distillation, which combines OPD with off-policy supervision on high-quality ``golden'' data.
Intuitively, the golden data serves as an anchor: it maintains a fraction of complete, non-truncated, and non-repetitive trajectories
throughout training, and keeps the OPD objective tied to reasonable reasoning behavior by reducing the influence of degenerate on-policy rollouts.

In addition to the on-policy rollouts used to optimize the OPD objective, we maintain a fixed dataset $\mathcal{D}_{\mathrm{gold}}$ of input-output pairs $(x,y)$ with complete and high-quality chain-of-thought solutions.
At each training step, we sample a set of prompts $x$ and, for each $x$,
include both the on-policy rollout generated by the student and a golden solution $(x,y)$ from $\mathcal{D}_{\mathrm{gold}}$ in the same minibatch.
Thus the model simultaneously sees its own trajectory and a high-quality
target for the same problem, and we optimize a combined loss over this mixed
supervision.
\begin{equation}
\mathcal{L}_{\text{mix}}(\theta)
=
\mathcal{L}_{\mathrm{OPD}}(\theta)
+
\lambda_{\text{gold}}\,
\mathbb{E}_{(x,y)\sim \mathcal{D}_{\mathrm{gold}}}
\big[
\mathcal{L}_{\mathrm{SFT}}(\theta; x,y)
\big],
\label{eq:mixture-objective}
\end{equation}
where $\mathcal{L}_{\mathrm{SFT}}$ is a
standard supervised loss and
$\lambda_{\text{gold}}$ controls the weight of the golden data.

Recent self-distillation methods~\cite{hubotter2026reinforcement,zhao2026self} have also leveraged high-quality ``golden''
responses, but for a different purpose. In those approaches, golden responses are typically used to refine the teacher signal itself. As noted by \citet{kim2026does}, this can suppress the teacher's uncertainty during reasoning and hurt student performance on complex problems. By contrast, mixture distillation leaves the original teacher-derived OPD signal unchanged on on-policy rollouts. Golden data is used only through an auxiliary off-policy SFT term that stabilizes training, and thus does not introduce this issue.

From a distributional perspective, mixture distillation can be viewed as
training on a mixture of two state distributions: the on-policy distribution
induced by the current student, and a fixed off-policy distribution induced by
$\mathcal{D}_{\mathrm{gold}}$.
This prevents the OPD objective from being driven solely by
truncation-dominated rollouts: gradients are rebalanced toward
high-quality, non-truncated reasoning trajectories, and the mixed
supervision in turn steers the student to generate better on-policy
samples during training.

\subsection{KL-Regularized Mixture Distillation}
While mixture distillation modifies the training distribution by injecting
off-policy golden data, it does not directly control the magnitude of the
student policy updates at each step. In the standard OPD setup, once the student drifts toward long, repetitive trajectories, the reverse-KL advantages start assigning large positive signals exactly to these states. To control this drift, we add a reference-based divergence constraint. In this work, we adapt KL regularization term on the policy itself.

We introduce a reference policy $\pi_{\mathrm{ref}}$ (e.g., the initial student checkpoint) and penalize deviations from this reference at visited prefix states. For a prefix state $s_t$ at step $t$, we define
$
\mathrm{KL}(s_t)
=
D_{\mathrm{KL}}\big(
\pi_\theta(\cdot \mid s_t)
\;\|\;
\pi_{\mathrm{ref}}(\cdot \mid s_t)
\big)
$
as the per-prefix KL between the student and reference policies. The KL-regularized Mixture Distillation loss is then defined as
\begin{equation}
\mathcal{L}_{\text{\Algnameabbr{}}}(\theta)
=
\mathcal{L}_{\text{mix}}(\theta)
+
\beta_{\text{KL}}\,
\mathbb{E}_{s_t}\big[
\mathrm{KL}(s_t)
\big],
\label{eq:opd-kl-loss}
\end{equation}
where $\beta_{\mathrm{KL}}>0$ controls the strength of the regularization.

\section{Experiment}
In this section, we conduct experiments to answer the following research questions about \Algnameabbr{}:
\textbf{RQ1}:
How does \Algnameabbr{} affect LLM reasoning performance on mathematical
reasoning benchmarks?
\textbf{RQ2}:
How does \Algnameabbr{} mitigate truncation-repetition saturation in OPD
training?
\textbf{RQ3}:
What are the individual effects of mixture distillation and
KL regularization?

\subsection{Experiment Setup}
\textbf{Training.} 
We build training data from OpenR1-Math-220k\footnote{\url{https://huggingface.co/datasets/open-r1/OpenR1-Math-220k}},
following the filtering procedure in~\cite{yan2025learning}. Prompts are sourced from NuminaMath~1.5~\cite{numina_math_datasets} and paired with reasoning traces generated by DeepSeek-R1~\cite{guo2025deepseek}. Starting from the default 94k-prompt split, we filter out generations that exceed 8192 tokens or are marked incorrect by Math-Verify\footnote{\url{https://github.com/huggingface/Math-Verify}}, and result in 46k prompts and high-quality demonstrations.

\textbf{Evaluation.} 
We evaluate on six widely used mathematical reasoning benchmarks:
AIME 2024, AIME 2025, AMC~\cite{numina_math_datasets},
Minerva~\cite{lewkowycz2022solving}, OlympiadBench~\cite{he2024olympiadbench},
and MATH500~\cite{hendrycks2021measuring}.
For AIME 2024, AIME 2025, and AMC, whose test sets are small, we
report avg@32.
For Minerva, OlympiadBench, and MATH500, we report pass@1.
The sampling temperature is $0.6$ for testing.

\textbf{Implementation Details.} 
For OPD training, we use a rollout batch size of 64 and sample 4 on-policy
trajectories per prompt. When applying mixture distillation, each prompt in
the batch is additionally paired with one off-policy golden solution from the dataset. Following recent OPD practice~\cite{lu2025onpolicydistillation}, we first perform supervised fine-tuning on 33k samples from the filtered dataset, and then run OPD on the remaining 13k samples. We generate rollouts with a sampling temperature of $1.0$ and optimize the student with Adam using a learning rate of $1\times10^{-6}$.
All experiments are conducted on 4$\times$H200 GPUs.

\textbf{Baseline Methods.} We benchmark \Algnameabbr{} against the following baselines using Qwen2.5-Math-1.5B and Qwen2.5-Math-7B~\cite{yang2024qwen25mathtechnicalreportmathematical}. For SFT methods, the base model is fine-tuned on the full 46k dataset.
For RL methods, we include GRPO~\cite{shao2024deepseekmath}, SimpleRL-Zero~\cite{zeng2025simplerl}, Oat-Zero~\cite{liu2025understanding}, PRIME-Zero~\cite{cui2025process}, and OpenReasonerZero~\cite{hu2025open}. We also compare with standard OPD~\cite{lu2025onpolicydistillation}. In addition, we adopt the same 33k/13k split as above: the model is first supervised fine-tuned on 33k examples and then trained with OPD on the remaining 13k examples. More details are listed in Appendix~\ref{apdx:baseline}.

\begin{table*}[t]
    \caption{Accuracy on six mathematical benchmarks for Qwen2.5-Math-7B. The best performance on each benchmark is highlighted in \textbf{bold}. For OPD-based methods we use OpenThinkerV3 as teacher models.}
    \vspace{-0.1cm}
    \label{tab:updated-math-results}
    \begin{center}
    \begin{footnotesize}
    \begin{sc}
    \resizebox{0.97\textwidth}{!}{
    \begin{tabular}{lccccccc}
    \toprule
    \textbf{Model} & \textbf{Avg} & \textbf{MATH-500} & \textbf{Minerva} & \textbf{Olympiad} & \textbf{AMC} & \textbf{AIME24} & \textbf{AIME25} \\
    \midrule    
    Qwen2.5-Math-7B              & 19.1 & 43.6 & 7.4  & 15.6 & 31.3 & 11.5 & 4.9 \\
    Qwen2.5-Math-7B-Instruct      & 37.6 & 80.4 & 32.7 & 41.0 & 48.5 & 12.5 & 10.2 \\
    \midrule
    SimpleRL-Zero                 & 37.4 & 76.0 & 25.0 & 34.7 & 54.9 & \textbf{27.0} & 6.8 \\
    OpenReasoner-Zero             & 41.0 & 82.4 & 33.1 & 47.1 & 52.1 & 16.5 & 15.0 \\
    PRIME-Zero                    & 40.8 & 81.4 & 39.0 & 40.3 & 54.0 & 17.0 & 12.8 \\
    Oat-Zero                      & 43.8 & 78.0 & 34.6 & 43.4 & 61.2 & 33.4 & 11.9 \\
    \midrule\midrule
    SFT                           & 44.1 & 82.6 & 40.8 & 43.7 & 52.8 & 22.2 & 22.3 \\
    GRPO                          & 45.5 & 84.4 & 39.3 & 46.8 & \textbf{62.0} & \textbf{25.1} & 15.3 \\
    OPD                           & 43.8   & 80.0 & 37.9 & 47.5 & 53.4   & 21.7   & 22.2   \\
    \Algnameabbr{}                & \textbf{47.6} & \textbf{84.6} & \textbf{43.4} & \textbf{49.3} & 58.1 & 24.7 & \textbf{25.2} \\
    
    \bottomrule
    \end{tabular}
    }
    \end{sc}
    \end{footnotesize}
    \end{center}
    \end{table*}

\subsection{Reasoning performance (RQ1)}
We compare \Algnameabbr{} with supervised and RL baselines on six mathematical reasoning benchmarks, reporting accuracies in Table~\ref{tab:updated-math-results} and Table~\ref{tab:updated-math-results-1.5B} for both Qwen2.5-Math-1.5B and Qwen2.5-Math-7B backbones.
Across both model scales, SFT and GRPO substantially improve over the base models, but standard OPD fails to match these gains despite leveraging on-policy samples and dense token-level supervision.
For example, on the 7B backbone, OPD achieves 43.8\% average accuracy, trailing SFT (44.1\%) and GRPO (45.5\%). A similar trend appears on the 1.5B backbone.
The results suggest that training instability limits the effectiveness of OPD. \Algnameabbr{} addresses this issue by stabilizing OPD with mixture distillation and KL regularization, yielding consistent improvements across scales.
As shown in Table~\ref{tab:updated-math-results-1.5B}, it boosts average accuracy from 28.9\% to 36.1\% (+7.2) on the 1.5B backbone, achieving the best performance. On 7B backbone, \Algnameabbr{} reaches 47.6\% average accuracy, surpassing all other methods.

We further compare \Algnameabbr{} with recent RLVR approaches, such as SimpleRL-Zero, OpenReasoner-Zero, PRIME-Zero and Oat-zero, on Qwen2.5-Math-7B backbone, as shown in Table~\ref{tab:updated-math-results}. \Algnameabbr{} achieves the best average accuracy (47.6\%), outperforming the strong zero-style methods.
These results indicate that stabilizing on-policy distillation can surpass carefully engineered RLVR pipelines, providing a simple yet effective alternative for improving mathematical reasoning.

\begin{figure*}[t]
    \centering
    \includegraphics[width=\linewidth,height=1\textheight,keepaspectratio]
    {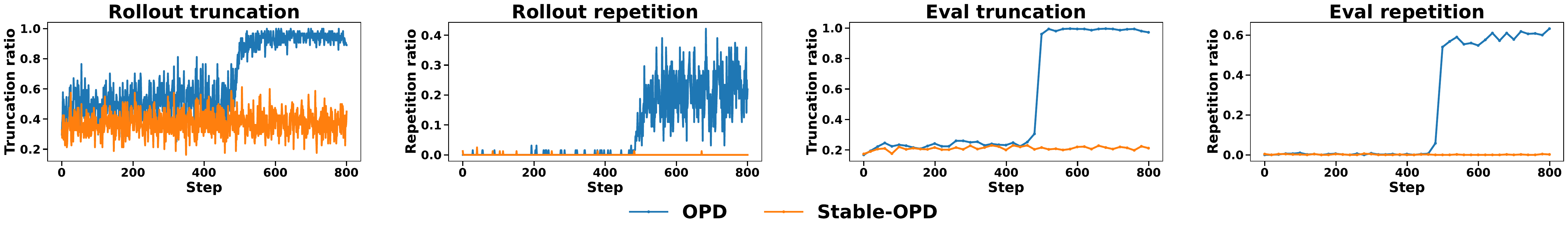}
    \vspace{-0.6cm}
    \caption{
    Training dynamics of OPD vs.\ \Algnameabbr{}. Student: Qwen2.5-Math-1.5B; Teacher: OpenThinker3-7B.
    }
    \vspace{-5mm}
    \label{fig:stable-opd-trunc-rep}
\end{figure*}

\subsection{Mitigating truncation-repetition inflation (RQ2)}
To assess how \Algnameabbr{} changes OPD training dynamics, we compare the same student-teacher settings and track rollout and evaluation truncation and repetition on MATH500 over training steps, as shown in Fig.~\ref{fig:stable-opd-trunc-rep} and Fig.~\ref{fig:stable-opd-trunc-rep-r1}.
Across both student-teacher groups, OPD exhibits the
truncation-repetition inflation: after an initially stable phase, rollout and evaluation truncation/repetition curves spike sharply and then remain at much higher levels than in the early stage, with only minor fluctuations. 
Under \Algnameabbr{}, the training dynamics remain stable.
For the Qwen2.5-Math-1.5B + OpenThinker3-7B setting, all four curves remain
flat throughout training: truncation ratios stay at moderate levels and
repetition ratios are near zero on both rollouts and evaluation prompts.
For the Qwen2.5-Math-1.5B + DeepSeek-R1-Distill-7B setting, we do observe a mild upward drift in truncation and repetition near the end of training, but the increase is much smaller than with OPD and occurs much later than the sharp inflation observed under OPD.
These results show that mixture distillation and KL regularization prevent OPD from abrupt truncation-repetition inflation regime.

\subsection{Ablation on mixture distillation and KL Regularization (RQ3)}
We ablate the two components of \Algnameabbr
{}, mixture distillation and KL
regularization, to quantify their 
contributions. We use DeepSeek-R1-Distill-7B 
as the teacher models. 
As shown in Table~\ref{tab:ablation}, compared to the base model, OPD improves 
performance to 28.0\%, but remains 
substantially below \Algnameabbr{} due to 
training instability. Adding KL 
regularization alone yields a modest but 
consistent gain (28.0$\rightarrow$29.7), 
indicating that constraining policy drift 
helps stabilize OPD updates. Combining KL regularization with mixture distillation brings a much larger improvement (29.7$\rightarrow$35.7), making it the strongest variant in the 1.5B setting.
This suggests that the two components are complementary: KL regularization limits abrupt policy shifts at 
the token-level, while mixture distillation 
provides a stable fraction of high-quality 
trajectories that anchors learning when 
on-policy rollouts start to degrade.
\begin{table}[t]
  \centering
  \caption{Ablation studies of KL regularization and mixture distillation on average accuracy.}
  \begin{tabular}{l c}
  \toprule
  Method & Avg. Acc (\%) \\
  \midrule
  Qwen2.5-MATH-1.5B & 16.0 \\
  OPD & 28.0 \\
  OPD + KL & 29.7 \\
  OPD + KL + Mixture Distillation & \textbf{35.7} \\
  \bottomrule
  \end{tabular}
  \vspace{-0.5cm}
  \label{tab:ablation}
  \end{table}

\section{Related Work}
\textbf{Length Bias in LLM Reasoning.}
Recent work such as Dr.GRPO~\cite{liu2025understanding} and DAPO~\cite{yu2025dapo} has observed that standard GRPO-style objectives implicitly favor longer responses, and proposes objective reweighting or normalization schemes to remove sample-length bias.
These methods focus on eliminating length-dependent gradient scaling at the sequence level, and have been shown to stabilize RL training under sparse rewards.
In contrast, our work identifies a qualitatively distinct training-time failure mode that arises in OPD. We show that repetitive continuations receive larger token-level advantages, and on-policy sampling amplifies their contribution once they become frequent, leading to abrupt repetition inflation and length explosion. This mechanism differs fundamentally from previously studied forms of length bias in RLVR and is specific to OPD dynamics.

\textbf{Knowledge Distillation}
Knowledge distillation (KD)~\cite{hinton2015distilling} is a widely used paradigm for model compression, where a student is trained under the guidance of a stronger teacher~\cite{rusu2015policy,gou2021knowledge}. In autoregressive generation, a common approach is to match the teacher’s conditional next-token distribution via token-level distillation, typically implemented as minimizing the forward KL divergence between student and teacher distributions at each decoding step~\cite{sanh2019distilbert}. An alternative is sequence-level distillation, where the student is trained on full sequences produced by the teacher~\cite{kim2016sequence}. While these objectives are stable and easy to optimize, they rely on a fixed supervision distribution, leading to train-inference mismatch once the student drifts from the supervised trajectories.

\textbf{On-policy Distillation.}
OPD trains the student on trajectories sampled from its current policy, while a teacher provides per-token guidance through KL-based regularization or closely related objectives~\cite{agarwal2024policy,lu2025onpolicydistillation,xiao2026mimo,yang2025qwen3,gu2023minillm}. By aligning the learning signal with the student’s own visitation distribution, these methods reduce the distribution shift that arises when supervision is collected off-policy. 
These works connect distillation to classic on-policy data aggregation in imitation learning, such as DAgger~\cite{ross2011reduction}, where an expert supplies corrective supervision on states encountered by the learner under its own policy.

\section{Conclusion}
We identify a failure mode of OPD characterized by abrupt rollout length inflation, truncation collapse, and repetition saturation. We show that this pathology arises from the interaction between student-induced data collection and the likelihood-based 
distillation objectives: repetitive tokens receive systematically larger advantages, and once sufficiently frequent, their disproportionate reward signal dominates gradient updates, creating a self-reinforcing feedback loop that implicitly favors increasingly long and repetitive rollouts.
To address this issue, we propose \Algnameabbr{}, a stabilized OPD framework that combines a reference-based divergence constraint with rollout mixture distillation. Across multiple mathematical reasoning benchmarks, \Algnameabbr{} consistently stabilizes OPD training and improves performance by 7.2\% on average compared to standard OPD baselines. 
\clearpage

\bibliographystyle{icml2026}
\bibliography{example_paper}

\appendix
\onecolumn

\section{Appendix}

\section{Use of LLMs}
We used a large language model only for spelling and grammar correction of the manuscript text. The LLM was not involved in research ideation, experimental design, data generation, analysis, or substantive writing beyond copy-editing. All content and claims were authored and verified by the authors, who take full responsibility for the paper. The LLM is not an author.

\section{Baseline Methods.}
\label{apdx:baseline}
We benchmark \Algnameabbr{} against the following baselines using Qwen2.5-Math-1.5B and Qwen2.5-Math-7B~\cite{yang2024qwen25mathtechnicalreportmathematical}. For SFT methods, the base model is fine-tuned on the full 46k dataset.
For RL methods, we include GRPO~\cite{shao2024deepseekmath}, trained on the
same dataset with verifiable rewards; SimpleRL-Zero~\cite{zeng2025simplerl}, which trains from Qwen2.5-Math-7B using rule-based reward; Oat-Zero~\cite{liu2025understanding} which trains from Qwen2.5-Math-7B and rule-based reward, proposing to remove the standard deviation in GRPO advantage computation and token-level normalization in policy loss computation; PRIME-Zero~\cite{cui2025process}, which uses policy rollouts and outcome labels through implict process rewards; and OpenReasonerZero~\cite{hu2025open} which is an open-source implementation of RLVR methods. We also compare with standard OPD~\cite{lu2025onpolicydistillation}, we adopt the same 33k/13k split as above: the model is first supervised fine-tuned on 33k examples and then trained with OPD on the remaining 13k examples.


\section{Additional Experiment Results}
\label{apx: additional exp}

\subsection{Additional Experiment Results on More Base models}
\label{apx:additional-exp-results-1.5B}
We also compare \Algnameabbr{} with supervised and RL baselines on six mathematical reasoning benchmarks for Qwen2.5-Math-1.5B backbone.
Similar to results in Table~\ref{tab:updated-math-results}, SFT and GRPO substantially improve over the base models, but standard OPD fails to match these gains despite leveraging on-policy samples and dense token-level supervision. The results suggest that training instability limits the effectiveness of OPD. \Algnameabbr{} addresses this issue by stabilizing OPD with mixture distillation and KL regularization, yielding consistent improvements across scales. On the 1.5B backbone, it boosts average accuracy from 28.9\% to 36.1\% (+7.2), achieving the best performance.

\begin{table*}[ht]
    \caption{Accuracy on six mathematical benchmarks for Qwen2.5-Math-1.5B. The best performance on each benchmark is highlighted in \textbf{bold}. For OPD-based methods we use OpenThinkerV3 as teacher models.}
    \vspace{-0.1cm}
    \label{tab:updated-math-results-1.5B}
    \begin{center}
    \begin{footnotesize}
    \begin{sc}
    \resizebox{0.97\textwidth}{!}{
    \begin{tabular}{lccccccc}
    \toprule
    \textbf{Model} & \textbf{Avg} & \textbf{MATH-500} & \textbf{Minerva} & \textbf{Olympiad} & \textbf{AMC} & \textbf{AIME24} & \textbf{AIME25} \\
    \midrule
    
    Qwen2.5-Math-1.5B          & 16.0 & 28.0 & 9.6 & 21.2 & 26.4 & 7.2 & 3.6 \\
    Qwen2.5-Math-1.5B-Instruct & 35.7 & 77.4 & 28.7 & 39.1 & 48.1 & 12.1 & 8.9 \\
    \midrule\midrule
    SFT                        & 31.9 & 70.6 & 26.8 & 31.3 & 37.8 & 11.7 & 13.2 \\
    GRPO                       & 30.1 & 61.8 & 26.8 & 32.0 & 40.2 & 11.8 & 7.7 \\
    OPD            & 28.9 & 56.7 & 23.4 & 31.0 & 35.9 & 11.1 & 15.0 \\
    \Algnameabbr{} & \textbf{36.1} & \textbf{73.9} & \textbf{32.6} & \textbf{37.4} & \textbf{43.0} & \textbf{13.8} & \textbf{16.0} \\    
    \bottomrule
    \end{tabular}
    }
    \end{sc}
    \end{footnotesize}
    \end{center}
    \end{table*}
    
\subsection{Additional Experiment Results on More Teacher models}
\label{apx: additional exp results for more teacher models}

To further validate the robustness of \Algnameabbr{} under different supervision sources, we conduct experiments with two teachers: DeepSeek-R1-Distill-7B and OpenThinkerV3. Table~\ref{tab:updated-math-results} reports the accuracy on six math benchmarks for all methods. Overall, \Algnameabbr{} with OpenThinkerV3 achieves the best average performance (36.1\%) and yields the strongest results on MATH-500 (73.9\%) and Olympiad (37.4\%). In contrast, \Algnameabbr{} distilled from the R1-distilled teacher performs best on Minerva (32.7\%), AIME24 (14.6\%), and AIME25 (17.2\%). Comparing teachers for the same student, distillation from the stronger OpenThinkerV3 teacher improves the average accuracy over R1-Distill-7B, aligning with the intuition that higher-quality teachers provide more informative token-level supervision.

\begin{table*}[ht]
\caption{Accuracy on six mathematical benchmarks. The best performance on each benchmark is highlighted in \textbf{bold}. \rionedistill denotes models distilled from DeepSeek-R1-Distill-7B, and \openthinker denotes models distilled from OpenThinkerV3.}
\label{tab:updated-math-results-add}
\begin{center}
\begin{small}
\begin{sc}
\resizebox{0.97\textwidth}{!}{
\begin{tabular}{lccccccc}
\toprule
\textbf{Model} & \textbf{Avg} & \textbf{MATH-500} & \textbf{Minerva} & \textbf{Olympiad} & \textbf{AMC} & \textbf{AIME24} & \textbf{AIME25} \\
\midrule
\midrule
Qwen2.5-Math-1.5B          & 16.0 & 28.0 & 9.6 & 21.2 & 26.4 & 7.2 & 3.6 \\
Qwen2.5-Math-1.5B-Instruct & 35.7 & 77.4 & 28.7 & 39.1 & 48.1 & 12.1 & 8.9 \\
\midrule\midrule
SFT                        & 31.9 & 70.6 & 26.8 & 31.3 & 37.8 & 11.7 & 13.2 \\
GRPO                       & 30.1 & 61.8 & 26.8 & 32.0 & 40.2 & 11.8 & 7.7 \\
OPD\rionedistill           & 28.0 & 58.4 & 22.4 & 26.9 & 36.2 & 10.9 & 13.1 \\
\Algnameabbr{}\rionedistill& 35.7 & 72.0 & \textbf{32.7} & 34.9 & \textbf{43.0} & \textbf{14.6} & \textbf{17.2} \\
\midrule
OPD\openthinker            & 28.9 & 56.7 & 23.4 & 31.0 & 35.9 & 11.1 & 15.0 \\
\Algnameabbr{}\openthinker & \textbf{36.1} & \textbf{73.9} & \textbf{32.6} & \textbf{37.4} & \textbf{43.0} & \textbf{13.8} & \textbf{16.0} \\
\bottomrule
\end{tabular}
}
\end{sc}
\end{small}
\end{center}
\end{table*}

\begin{figure*}[ht]
    \centering
    \includegraphics[width=\linewidth]{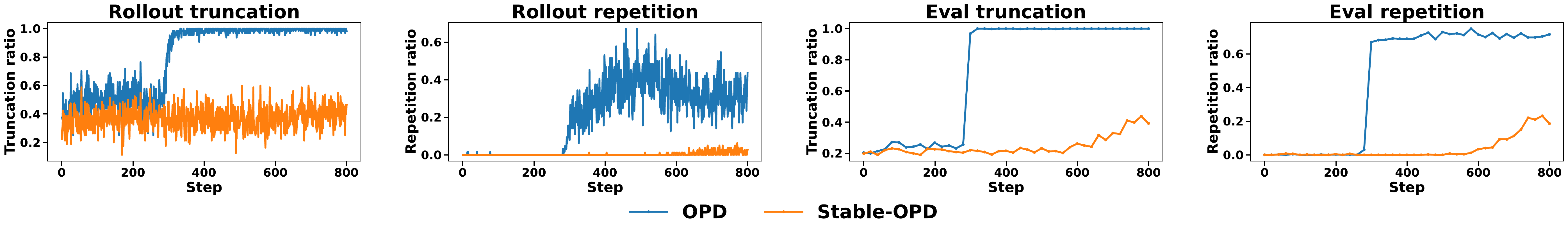}
    \caption{
    Training dynamics of OPD vs.\ \Algnameabbr{}. Student: Qwen2.5-Math-1.5B; Teacher: DeepSeek-R1-Distill-7B.
    }
    \vspace{-6mm}
    \label{fig:stable-opd-trunc-rep-r1}
\end{figure*}

\section{More Dynamics Analysis of OPD}
\begin{figure}[h!]
  \centering

  \begin{subfigure}[t]{0.5\textwidth}
      \centering
      \includegraphics[width=\linewidth]{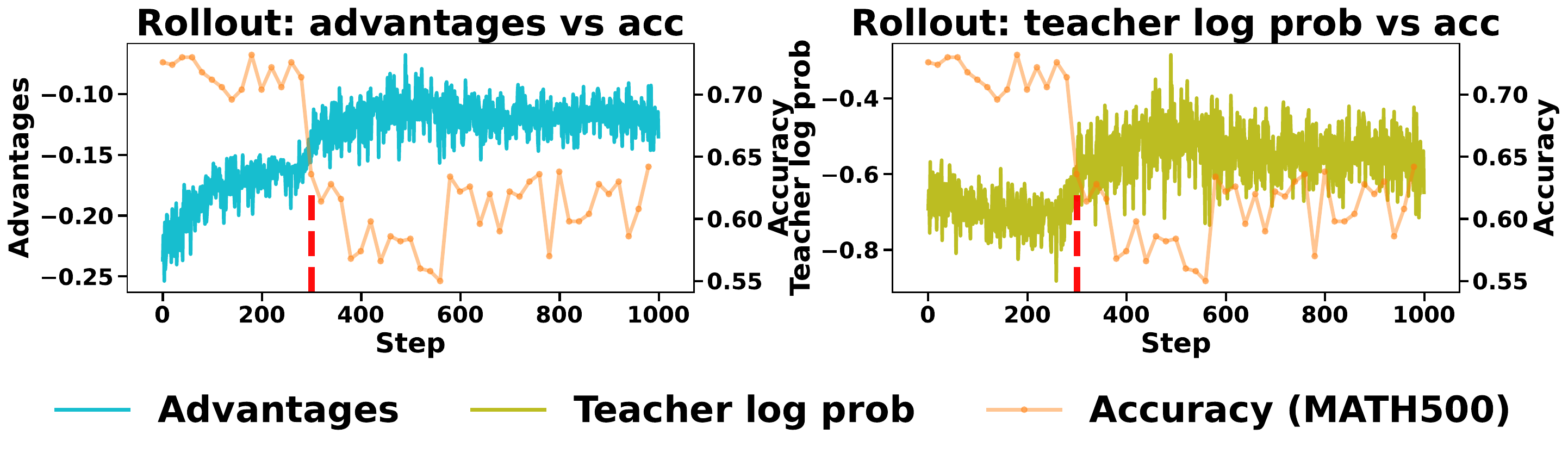}
      \vspace{-6mm}
      \caption{\footnotesize Student: Qwen2.5-Math-1.5B; Teacher: DeepSeek-R1-Distill-7B}
      \vspace{1mm}
  \end{subfigure}
  \begin{subfigure}[t]{0.5\textwidth}
      \centering
      \includegraphics[width=\linewidth]{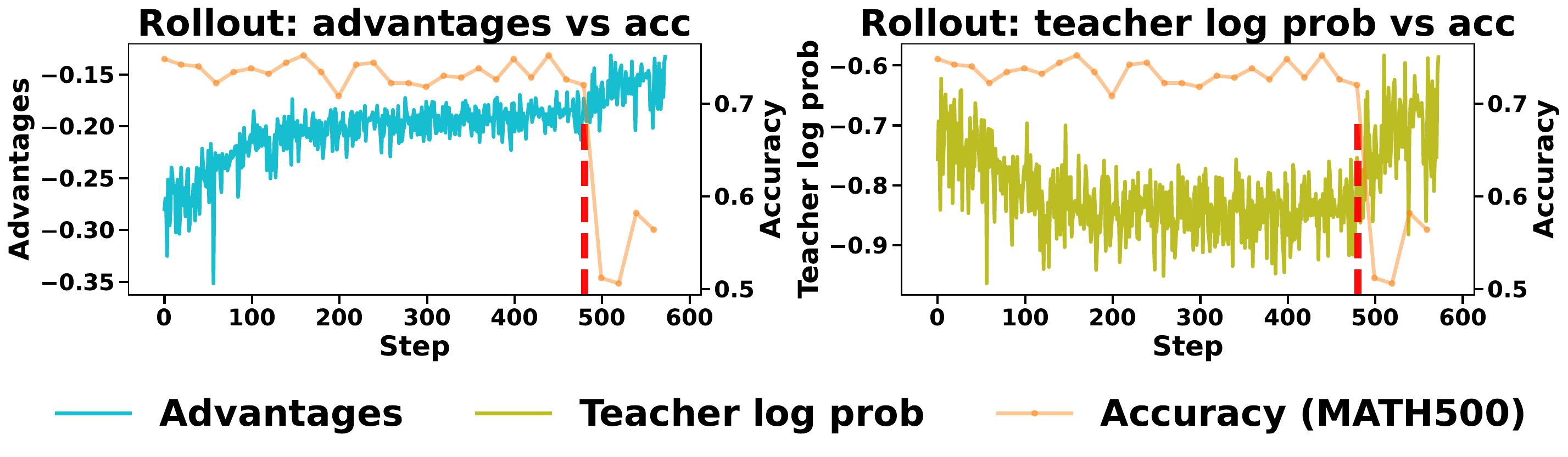}
      \vspace{-6mm}
      \caption{\footnotesize Student: Qwen2.5-Math-1.5B; Teacher: OpenThinker3-7B}
      \vspace{1mm}
  \end{subfigure}
  \begin{subfigure}[t]{0.5\textwidth}
      \centering
      \includegraphics[width=\linewidth]{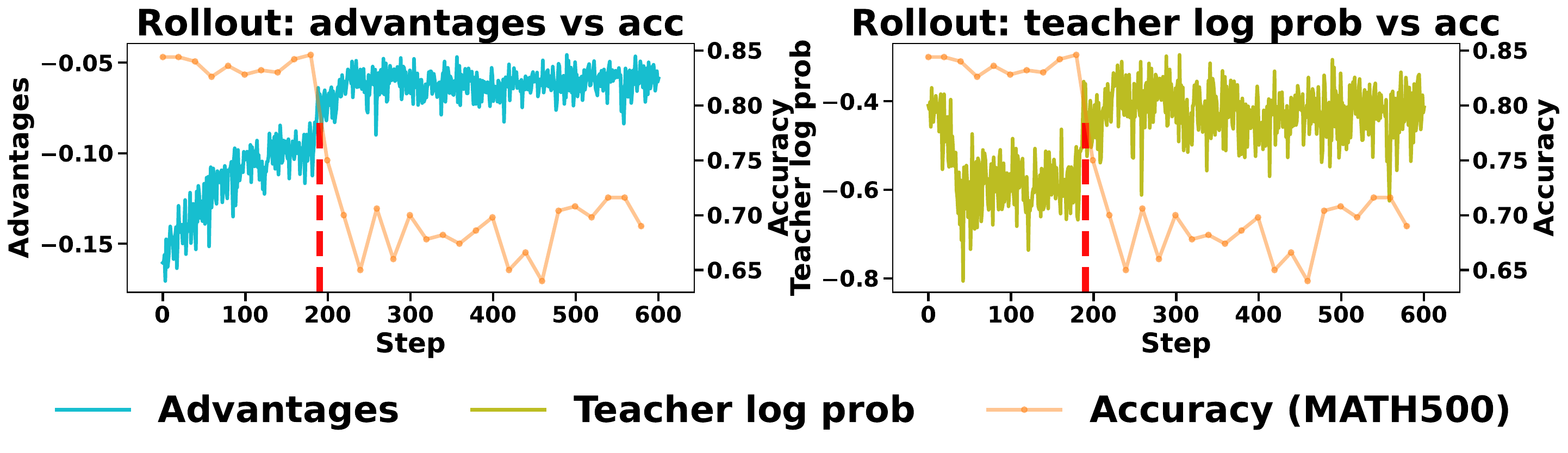}
      \vspace{-6mm}
      \caption{\footnotesize Student: Qwen2.5-Math-7B; Teacher: DeepSeek-R1-Distill-7B}
      \vspace{1mm}
  \end{subfigure}
  \caption{
  Dynamics of OPD across training.
  Each panel shows truncation/repetition for both rollout and evaluation versus accuracy.
  Sudden accuracy changes often align with abrupt shifts in teacher log-probabilities and advantage estimates.
  }
  \label{fig:trunc_rep_acc_panels_singlecol}
\end{figure}

As shown in Fig.~\ref{fig:trunc_rep_acc_panels_singlecol}, abrupt accuracy changes are often synchronized with sharp shifts in the underlying training signals, especially the advantage and the teacher log-probabilities. These synchronized spikes often co-occur with increased truncation or repetition, suggesting a strong correlation between unstable teacher-guided supervision and sudden performance fluctuations.

\end{document}